\documentclass[11pt]{article}
\usepackage{amsmath}
\usepackage{amsfonts}
\usepackage[colorlinks=true, linkcolor=blue, citecolor=blue, urlcolor=blue]{hyperref}
\usepackage{marvosym}  

\usepackage{amssymb}
\usepackage[final]{acl}
\usepackage{booktabs}
\usepackage{multirow}
\usepackage{times}
\usepackage{latexsym}
\usepackage{algorithm}
\usepackage{algpseudocode}
\usepackage[T1]{fontenc}
\usepackage{tcolorbox}
\usepackage{float}
\usepackage{colortbl}
\usepackage{fontawesome5}  

\usepackage[utf8]{inputenc}

\usepackage{microtype}

\usepackage{inconsolata}

\usepackage{graphicx}
\usepackage{enumitem}
\usepackage{xcolor}
\usepackage{tcolorbox}
\usepackage{colortbl}
\newcommand{\reason}[1]{\textcolor{blue}{\texttt{<reasoning>}} #1 \textcolor{blue}{\texttt{</reasoning>}}}
\newcommand{\search}[1]{\textcolor{cyan}{\texttt{<search>}} #1 \textcolor{cyan}{\texttt{</search>}}}
\newcommand{\tool}[1]{\textcolor{cyan}{\texttt{<tool\_call>}} #1 \textcolor{cyan}{\texttt{</tool\_call>}}}
\newcommand{\info}[1]{\textcolor[RGB]{185,154,58}{\texttt{<information>}} #1 \textcolor[RGB]{185,154,58}{\texttt{</information>}}}
\newcommand{\answer}[1]{\textcolor[RGB]{0,100,0}{\texttt{<answer>}} #1 \textcolor[RGB]{0,100,0}{\texttt{</answer>}}}
\newcommand{\finaltools}[1]{\textcolor[RGB]{0,100,0}{\texttt{<final\_tools>}} #1 \textcolor[RGB]{0,100,0}{\texttt{</final\_tools>}}}
\definecolor{myblue}{RGB}{0, 76, 153}
\definecolor{myred}{RGB}{197, 0, 0}
\definecolor{mygreen}{RGB}{0, 128, 0}
\definecolor{bg_gray}{RGB}{245, 245, 245}
\title{ToolOmni: Enabling Open-World Tool Use via Agentic learning with Proactive Retrieval and Grounded Execution}

\author{Shouzheng Huang,  Meishan Zhang, \textbf{Baotian Hu\textsuperscript{\Letter}\thanks{\textsuperscript{\Letter}Corresponding author.},  Min Zhang} \\ 
        Harbin Institute of Technology (Shenzhen) \\
        \texttt{huangshouzheng@stu.hit.edu.cn,} \texttt{mason.zms@gmail.com,}\\ 
        \texttt{\{hubaotian, zhangmin2021\}@hit.edu.cn} \\
        \vspace{0.2cm} 
  \href{https://github.com/Huangsz2021/ToolOmni}{\faGithub~\texttt{https://github.com/Huangsz2021/ToolOmni}}
  }
\makeatletter
\def\thanks#1{\protected@xdef\@thanks{\@thanks
        \protect\footnotetext{#1}}}
\makeatother

\begin{document}
\maketitle
\begin{abstract}

%
Large Language Models (LLMs) enhance their problem-solving capability by utilizing external tools. However, in \textbf{open-world} scenarios with massive and evolving tool repositories, existing methods relying on static embedding retrieval or parameter memorization of tools struggle to align user intent with tool semantics or generalize to unseen tools, respectively, leading to suboptimal accuracy of open-world tool retrieval and execution.
%
To address these, we present \textbf{ToolOmni}, a unified agentic framework that enables LLMs for open-world tool use by proactive retrieval and grounded execution within a reasoning loop.
First, we construct a cold-start multi-turn interaction dataset to instill foundational agentic capabilities via Supervised Fine-Tuning (SFT).
%
%
Then, we introduce open-world tool learning based on a \textbf{Decoupled Multi-Objective GRPO} algorithm, which simultaneously optimizes LLMs for both tool retrieval accuracy and execution efficacy in online environments. 
%
%
%
Extensive experiments demonstrate that ToolOmni achieves state-of-the-art performance both in retrieval and execution, surpassing strong baselines by a significant margin of \textbf{+10.8\%} in end-to-end execution success rate, while exhibiting exceptional robustness and generalization capabilities.
\end{abstract}

%
\section{Introduction}
\begin{figure}[t]
    \centering
    \includegraphics[width=0.48\textwidth]{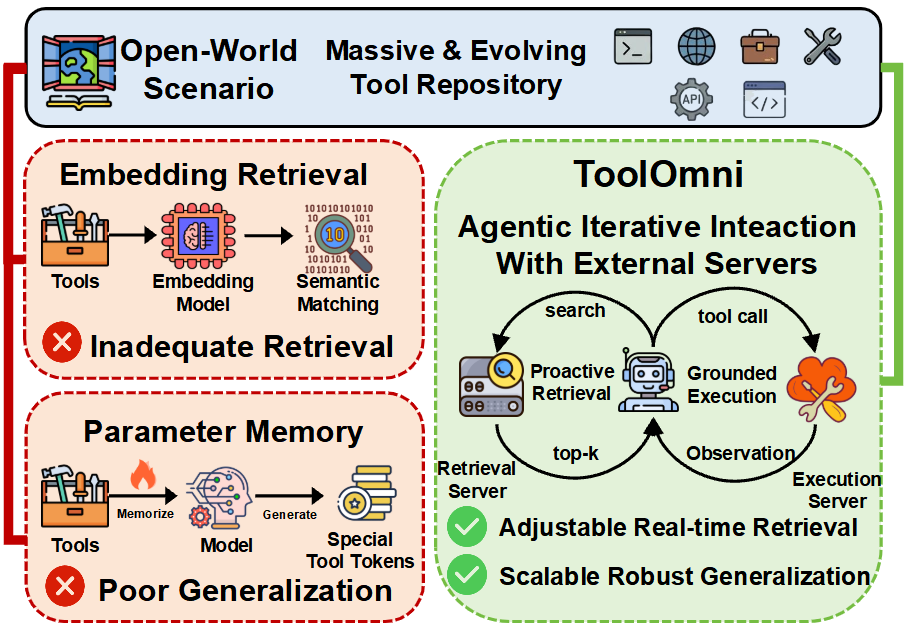} 
    \caption{Motivation for \textbf{ToolOmni} in Open-World Scenarios:
    Embedding retrieval methods struggle with \textit{Massive tools}, often resulting in low retrieval accuracy due to shallow matching; 
    Parameter memory methods fail to adapt to \textit{Evolving tools}, suffering from poor generalization to unseen tools. \
    \textbf{ToolOmni} overcomes these limitations via a unified agentic framework that couples \textbf{Proactive Retrieval} with \textbf{Grounded Execution} , enabling effective open-world tool use.}
    \label{fig:motivation}
    \vspace{-15pt} 
\end{figure}
Tool Learning with LLM achieves higher accuracy, efficiency, and autonomy in problem solving by combining the strengths of specialized tools and foundational models \cite{nakano2021webgpt,yao2022webshop,DBLP:conf/iclr/YaoZYDSN023,DBLP:conf/nips/SchickDDRLHZCS23,DBLP:conf/iclr/QinLYZYLLCTQZHT24,wu2023visual,li2025perception}.
Efforts in this field have predominantly focused on teaching models to effectively use tools via demonstration-based learning, typically leveraging Supervised Fine-Tuning(SFT) on curated expert trajectories \cite{nakano2021webgpt,yao2022webshop,DBLP:conf/nips/SchickDDRLHZCS23,li2025uni}, or feedback-based learning, which aligns model behavior with feedback from both environment and human through Reinforcement Learning \cite{schulman2017proximal,christiano2017deep,nakano2021webgpt,baker2022video,li2025torl,qian2025toolrl}.
However, in \textbf{open-world} scenarios characterized by massive and dynamically updated tool repositories, models must not only understand how to use tools but also master the ability to search and select the correct ones.
%
%

As shown in Fig.\ref{fig:motivation},
to help narrow the scope of relevant tools, prevailing solutions typically adopt a pipeline approach, employing embedding models to retrieve relevant tools based on query similarity before passing them to the execution agent \cite{DBLP:conf/iclr/QinLYZYLLCTQZHT24,DBLP:conf/emnlp/ChenYSWCBLP24,xu2024enhancing}.
However, this paradigm operates passively, which decouples the retrieval process from the agent's reasoning, preventing the execution model from proactively participating in tool selection or refining the search based on task-specific needs.
Consequently, these methods often struggle to effectively \textbf{align user intent} with the functionally essential tools in complex scenarios.
Alternatively, some approaches~\cite{DBLP:conf/iclr/WangHJWB025,DBLP:conf/nips/SchickDDRLHZCS23,su2025toolscaler} fine-tune models to internalize tool documentation into parametric knowledge.
While effective, this paradigm requires expensive retraining whenever the toolset updates, severely limiting \textbf{generalizability in dynamic environments}.
Recently, there has been a growing interest in agentic training frameworks that leverage Reinforcement Learning with Verifiable Rewards(RLVR) to unify reasoning with active environment interaction.
Building on algorithms like Group Relative Policy Optimization(GRPO)~\cite{shao2024deepseekmath} and Proximal Policy Optimization(PPO)~\cite{schulman2017proximal}, 
recent works \cite{jin2025search,xue2025simpletir,qian2025toolrl,li2025torl} have demonstrated that LLMs can be trained to iteratively interact with the external environment, effectively invoking tools and leveraging feedback to enhance both performance and generalization.
Nevertheless, these works often restrict LLMs to a limited toolset such as search engines and code executors, which constrains their applicability to diverse and dynamic open-world scenarios.

%
To address the above issues, we propose a unified agentic framework \textbf{ToolOmni} that integrates proactive retrieval and grounded execution into a unified end-to-end process, scaling agentic capabilities to dynamic, open-world tool scenarios.
%
%
%
To build ToolOmni, we first conduct \textbf{Tool Learning cold start} phase with a high-quality hybrid dataset that integrates both retrieval and execution trajectories. 
This stage enables the model to acquire the foundational capabilities required for effective tool interaction.
Building on this foundation, the second stage adopts an \textbf{Open World Tool Learning} process based on an enhanced Group Relative Policy Optimization algorithm.
Unlike naive GRPO, which relies on a single reward, our framework treats retrieval and execution as interconnected yet distinct sub-tasks, where we compute task-specific rewards and advantages for retrieval and execution independently, and integrate them into a single optimization, enabling the synchronized optimization of both capabilities.
This is crucial because decoupling allows us to provide finer-grained process supervision, ensuring that both retrieval recall and execution reasoning are optimized precisely without mutual interference.

To evaluate the effectiveness of ToolOmni, we conduct extensive experiments on the ToolBench benchmark.
The results demonstrate that ToolOmni achieves superior performance in both tool retrieval and task execution, particularly in open-world scenarios with massive candidate tools\textbf{(+11.9\%)}.
Additionally, ToolOmni exhibits exceptional robustness to unseen instructions and tools, demonstrating that it learns universal tool-use mechanisms rather than relying on rigid memorization.
The contributions of this work can be summarized as follows:
\begin{itemize}[leftmargin=*]
    \item We introduce \textbf{ToolOmni}, an end-to-end tool agentic framework that integrates proactive tool retrieval with grounded execution within a unified reasoning loop.
    %
    \item We propose a two-stage training strategy that integrates a supervised cold-start for foundational tool retrieval and execution with GRPO-based RL for the subsequent synchronized optimization.
    %
    \item Extensive experiments demonstrate that \textbf{ToolOmni} not only achieves superior performance in both tool retrieval and execution, but also shows strong robustness and generalizability to unseen domains.
\end{itemize}

\section{Related Works}
\subsection{Tool Retrieval in Open-World Scenarios}
In open-world scenarios, a common approach for tool retrieval is to employ an embedding model~\cite{shi2025retrieval,robertson2009probabilistic,DBLP:conf/iclr/QinLYZYLLCTQZHT24,qu2024towards,zhao2025kalm,zhao2026lmeb} that retrieves top-k relevant tools based on semantic similarity, narrowing the scope of candidate tools.
Some works~\cite{DBLP:conf/emnlp/ChenYSWCBLP24,xu2024enhancing,zheng2024toolrerank,kachuee2025improving} improve retrieval performance by using LLM to rewrite queries or expand tool documentation.
Alternatively, another method trains the LLM to encode tool information into its parametric knowledge, enabling the model to directly generate corresponding tool identifiers to accomplish retrieval~\cite{DBLP:conf/iclr/WangHJWB025,su2025toolscaler}.
Our approach also utilizes embedding-based retrieval, yet transforms the interaction paradigm where the agent proactively formulates queries and invokes the embedding model as an executable tool.
\subsection{LLM Tool Execution}
Numerous studies have focused on augmenting LLMs with external tools to enhance their specialization and efficiency in solving complex tasks~\cite{liang2024taskmatrix,xu2023tool,DBLP:conf/nips/SchickDDRLHZCS23,yao2022webshop,nakano2021webgpt}.
ReAct~\cite{DBLP:conf/iclr/YaoZYDSN023} establishes a Thought-Action-Observation loop, where the model generates explicit reasoning traces to justify and orchestrate tool execution in an interleaved manner.
ToolLLM~\cite{DBLP:conf/iclr/QinLYZYLLCTQZHT24} adopts a tree search framework (DFSDT), allowing the model to explore multiple execution paths and back-track based on tool feedback to solve multi-step tasks.
Meta-Tool~\cite{DBLP:conf/acl/QinZM0025} incorporates a plug-and-play retrieval module allowing the model to proactively search for relevant APIs, yet it treats retrieval and execution as isolated pipeline stages without joint optimization.
More recently, agentic training frameworks that leverage RLVR have been employed to enhance the LLM's ability to use external tools~\cite{jin2025search,xue2025simpletir,qian2025toolrl,li2025torl,li2025deepagent}.
%
These studies treat external search and code execution as executable tools, facilitating an agentic, end-to-end reasoning paradigm for complex task solving.
Inspired by them, ToolOmni broadens this scope to open-world by unifying proactive tool discovery and execution into an end-to-end process for complex task resolution.

\section{Methodology}

\begin{figure*}[t]
    \centering
    \includegraphics[width=0.95\textwidth]{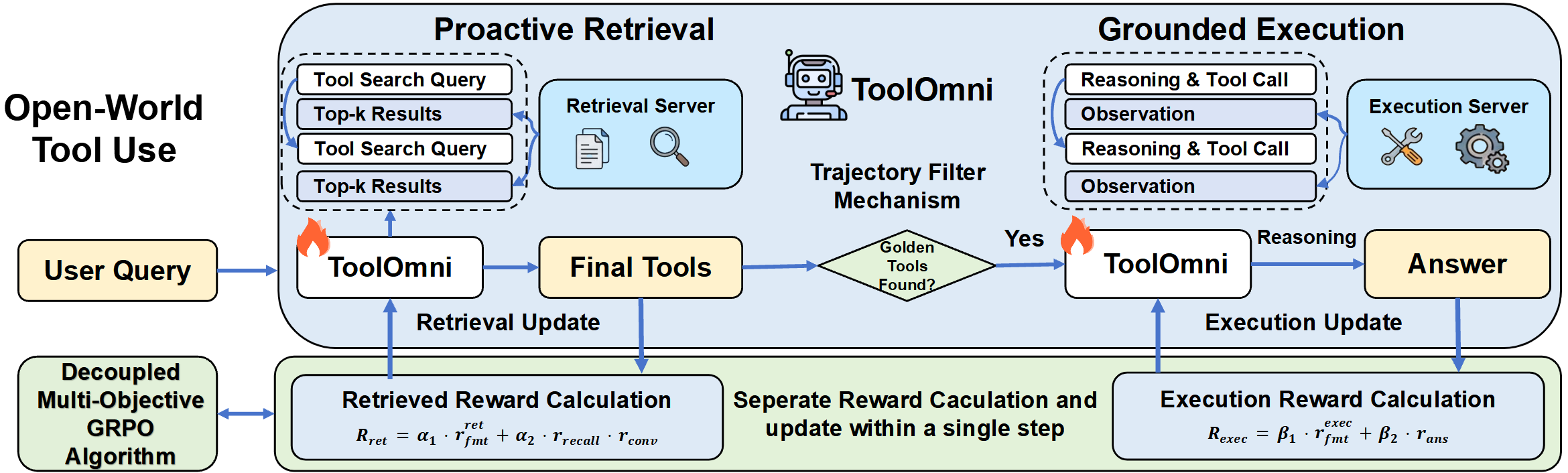} 
    \caption{Overview of the \textbf{ToolOmni} framework. The pipeline operates in two decoupled phases: 
\textbf{Proactive Retrieval}: The agent iteratively interacts with the retrieval server to curate a candidate tool set. 
%
%
\textbf{Grounded Execution}: With retrieval results, the agent performs reasoning and tool invocation to generate the final answer.}
\label{fig:method_framework}
\vspace{-5pt} 
\end{figure*}

\subsection{Problem Formulation}
The goal of open-world tool-use agents is to address a user query $Q$ by interacting with a large-scale tool repository~$\mathcal{T} = \{t_1, t_2, \dots, t_N\}$, where $N$ is the number of candidate tools. 
Given the massive scale of $\mathcal{T}$ (e.g., $N > 10,000$), ToolOmni models the open-world tool-use process as a cascaded retrieval-execution framework driven by the policy $\pi_{\theta}$.
At each turn $t$, the agent first performs an iterative \textit{Proactive Retrieval} phase to identify a task-complete set of tools, followed by a \textit{Grounded Execution} phase to invoke them. 
The trajectory sequence $\tau$ and final answer $a$ can be formulated as follows:
\begin{align}
\tau &= \big\{ 
    \underbrace{(r_t^{ret}, \alpha_t^{ret}, \mathcal{T}_{sub})}_{\text{Proactive Retrieval}}, \;
    \underbrace{(r_t^{exe}, \alpha_t^{exe}, o_t)}_{\text{Grounded Execution}}
\big\}_{t=1}^T, \notag \\
a &\sim \pi_\theta(Q, \tau)
\end{align}
where the cascaded retrieval-execution framework at each turn is determined by the policy:
%
\[
\tau =
\begin{cases} 
(r_{t}^{ret}, \alpha_{t}^{ret}) \sim \pi_\theta(\cdot \mid s_t, Q), & \text{Ret.} \\
(r_{t}^{exe}, \alpha_{t}^{exe}) \sim \pi_\theta(\cdot \mid s_t, Q, \mathcal{T}_{sub}), & \text{Exec.}
\end{cases}
\]
At each turn $t$, the agent's state $s_t$ consists of the history of all previous actions and the corresponding observations, i.e., $s_t=(a_1,s_1,...,a_{t-1},o_{t-1})$.
As the Fig.\ref{fig:method_framework} shows, the agent iteratively interleaves reasoning $r_i^{ret}$ and actions $\alpha_i^{ret}$ in the Proactive Retrieval phase, 
where each action $\alpha_i^{ret}$ either queries $q_i$ to retrieve relevant tools or finalizes the task-complete toolset $\mathcal{T}_{sub}$.
After finalizing the toolset $\mathcal{T}_{sub}$, the agent enters the Grounded Execution phase, where it interleaves reasoning $r_t^{exe}$ and tool-call actions $\alpha_t^{exe}$ grounded in the retrieved tool documentation to generate observations $o_t$ or the final answer $a$.

\subsection{Cold-start Tool Learning}
To endow ToolOmni with basic tool-use capabilities, we first perform a SFT phase, serving as a cold-start initialization for the model.

\paragraph{Data Curation.} 
Our training dataset is derived from ToolBench~\cite{DBLP:conf/iclr/QinLYZYLLCTQZHT24} and carefully curated to support both retrieval and execution phases.
For tool retrieval, we first select a subset of 80,000 queries to train a specialized retrieval-only model. 
Using this model, we perform rejection sampling to remove low-quality instances, resulting in a high-quality corpus of approximately 28,000 retrieval trajectories.
For tool execution, we extract around 33,000 trajectories from the ToolBench training set, including both correct and incorrect execution paths. 
To ensure data quality, we employ Qwen-2.5-32B as an automated judge to rigorously validate each trajectory.
Further details are presented in Appendix\ref{sec:appendix_data}.

\paragraph{SFT Objective.}
Given the dataset of high-quality trajectories $\mathcal{D} = \{(\tau_{ret}, \tau_{exe})\}$, we optimize the model using the standard cross-entropy loss:
\begin{equation}
    \mathcal{L}_{SFT} = -\sum_{(x,y) \in \mathcal{D}} \log \pi_{\theta}(y \mid x)
\end{equation}
%
Following prior works~\cite{jin2025search,huang2025reinforced,jin2025empirical,sun2505zerosearch}, we compute token-level losses exclusively on the agent-generated reasoning traces $r$ and actions $\alpha$, masking all external observations $o$.
This strategy prevents the model from predicting environmental dynamics, thereby stabilizing the policy gradient optimization.

%
%
%

\subsection{Open-world Tool Learning}
While SFT establishes basic tool-use capabilities, it relies on imitating high-quality labeled trajectories, 
restricting exploration of diverse trajectories and limiting scalability to large-scale, open-world tool scenarios.
To bridge this gap, we introduce open-world tool learning based on GRPO, which enables the agent to optimize its proactive retrieval and grounded execution actions through iterative trial and error.

\subsubsection{Proactive Tool Retrieval}

The agentic interaction begins with the proactive tool retrieval phase. Unlike existing tool learning approaches that rely on a single-turn passive retrieval, ToolOmni autonomously determines whether and what to retrieve. 
Formally, given the user instruction $Q$, the policy $\pi_\theta$ analyzes the user's intents and then formulates a tool search query $q_t$ encapsulated within special tags: \search{$q_t$}.
Upon receiving $q_t$, the retrieval server encodes it using a pre-trained embedding model $E(\cdot)$ and then performs retrieval. 
It computes the cosine similarity between the query vector $E_{q_t}$ and the pre-indexed tool embeddings, returning the top-$k$ candidates:
\begin{equation}
    \mathcal{T}_{ret}^t = \underset{\tau \in \mathcal{T}}{\operatorname{top-k}} \left( \operatorname{cos}(E_{\tau}, E_{q_t}) \right),
\end{equation}
where $\operatorname{cos}(\cdot)$ denotes cosine similarity function.
ToolOmni iterates proactive retrieval, autonomously generating multiple search queries as needed with real-time retrieval.
Once a complete set of tools sufficient for the task has been collected, ToolOmni selects useful tools from the candidates and arranges them into a sorted subset that is finalized within \tool{and} tags.
\begin{equation}
    \mathcal{T}_{sub} = \pi_\theta(P_{ret},Q,\bigcup_{t} \mathcal{T}_{ret}^t),
\end{equation}

\subsubsection{Grounded Tool Execution}
\label{sec:execution}
After obtaining the sorted tool subset $\mathcal{T}_{sub}$, ToolOmni is instructed via the execution prompt $P{exec}$ to enter the execution phase.
We employ a strategic trajectory filtering procedure to ensure both the training stability of the execution policy and the solvability of the queries.
Specifically, we only retain trajectories where the generated subset $\mathcal{T}_{sub}$ successfully recall all ground-truth tools ($\mathcal{T}_{gold} \subseteq \mathcal{T}_{sub}$).
Based on these valid retrieval results, the policy $\pi_\theta$ then conducts multi-turn interleaved reasoning and tool invocation to solve the user's query.
%

%
Specifically, at each step $t$, ToolOmni first conducts reasoning inside \reason{and} tags.
Subsequently, it invokes a tool by explicitly specifying the target function name and its required arguments, e.g., 
\tool{\{"tool\_name": "genderize","tool\_input": \{"name": "john"\}\}}.
%
To ensure stable and efficient online reinforcement learning, we deploy an \textit{LLM-based Tool Simulator} that emulates the environment's feedback~\cite{guo2024stabletoolbench}. 
The simulator produces realistic execution results for tool invocations, which are enclosed within \info{and} tags and appended to the ongoing context.
%
The iterative cycle of reasoning and tool invocation continues until it derives the final solution, which is ultimately presented within \answer{and} tags.
Finally, the grounded tool execution process can be formulated as:
\begin{equation}
    y = \pi_\theta(P_{exec}, Q, \mathcal{T}_{sub}, O_{tool}).
\end{equation}
Given the open-ended nature of tool execution, rigid rule-based matching is insufficient to verify the final answer. 
Instead, we utilize a reward model to evaluate each trajectory holistically. A positive reward is assigned when the agent successfully invokes the required tools and produces a correct final answer.
\subsubsection{Open-world Tool Use Rewards}
In the RL process, the reward functions are utilized to steer the model towards desired properties. 
Below, we design retrieval and execution rewards to induce the model to perform effective proactive retrieval and reliable grounded execution.
The retrieval reward $R_{ret}$ comprises three weighted components: ensuring format correctness, maximizing recall of ground-truth tools, and promoting effective conversion of retrieved information:
\begin{equation}
    R_{ret} = \alpha_1 \cdot r_{fmt}^{ret} + \alpha_2 \cdot r_{rec} \cdot r_{conv}.
\end{equation}
Here, $r_{fmt}^{ret} \in \{0, 1\}$ ensures format compliance; $r_{rec} \in [0, 1]$ measures the recall of ground-truth tools $\mathcal{T}_{gold}$ within the cumulative retrieved set $\mathcal{T}_{ret}$; and $r_{conv} \in [0, 1]$ rewards the proportion of these retrieved tools that are ultimately selected into the final set, encouraging effective utilization.


The execution reward $R_{exec}$ is similarly formulated as a weighted combination of format and outcome correctness:
\begin{equation}
    R_{exec} = \beta_1 \cdot r_{fmt}^{exec} + \beta_2 \cdot r_{ans}.
\end{equation}
The term $r_{fmt}^{exec} \in \{0, 1\}$ enforces structural validity of the reasoning path and tool invocations, while $r_{ans} \in \{0, 1\}$ is a binary signal from the reward model indicating the correctness of the answer.

   
\subsubsection{Online Policy Optimization}

\paragraph{Group-Relative Advantage Estimation.}
For each query $x$, the agent generates a group of $G$ outputs $\{o_1, \dots, o_G\}$ sampled from the current policy $\pi_\theta$. We compute the advantages for retrieval and execution independently using the group-relative standard:
\begin{equation}
    A_{task}^{(i)} = \frac{R_{task}^{(i)} - \text{mean}(\{R_{task}\}_G)}{\text{std}(\{R_{task}\}_G) + \epsilon}
\end{equation}
where $task \in \{ret, exec\}$. By normalizing rewards within the group specifically for each sub-task, we isolate the learning signals, preventing the sparsity of execution rewards from destabilizing the retrieval learning process.
\paragraph{Decoupled Policy Gradient.}
Based on the estimated advantages, we optimize the policy $\pi_\theta$ by maximizing the surrogate objective. 
Consistent with the SFT stage, we apply token-level masking to ensure that gradients are back-propagated solely through the agent-generated tokens. 
The final optimization objective is formulated as:
\begin{equation}
\nonumber
\begin{split}
    \mathcal{J}_{GRPO}(\theta) = \mathbb{E} \Bigg[ \frac{1}{G} \sum_{i=1}^G \frac{1}{|y_i|} \sum_{t=1}^{|y_i|} \mathbb{I}(t \in \mathcal{M}_{task}) \\
    \min \left( \rho_{i,t}(\theta) \hat{A}_{i}, \text{clip}(\rho_{i,t}(\theta), 1-\epsilon, 1+\epsilon) \hat{A}_{i} \right) \Bigg]
\end{split}
\end{equation}
\paragraph{Optimization Stability.}
To further stabilize the training, we implement two key optimizations for GRPO in this context:
\textbf{(1)Separated Update}: Instead of summing the gradients directly, we perform the updates for retrieval and execution sequentially within a single step. This prevents gradient conflict where the magnitude of one objective overwhelms the other;
\textbf{(2)Selective Rollout}: As detailed in Sec. \ref{sec:execution}, we enforce a strict filtering mechanism where the execution generation is initiated only when the retrieval stage successfully recalls all golden tools ($\mathcal{T}_{gold} \subseteq \mathcal{T}_{ret}$). By excluding invalid retrieval instances prior to rollout, we ensure that the execution policy is trained exclusively on high-quality, grounded contexts.

\begin{table*}[t]
\centering
\small
\tabcolsep=0.2cm
\renewcommand\arraystretch{1.1}
\setlength{\tabcolsep}{5pt} 
\begin{tabular}{l|ccc|ccc|ccc|c}
\toprule
\multirow{2}{*}{\textbf{Method}} & \multicolumn{3}{c|}{\textbf{I1}} & \multicolumn{3}{c|}{\textbf{I2}} & \multicolumn{3}{c|}{\textbf{I3}} & \multirow{2}{*}{\textbf{Avg.}} \\
\cline{2-10}
 & \textbf{@1} & \textbf{@3} & \textbf{@5} & \textbf{@1} & \textbf{@3} & \textbf{@5} & \textbf{@1} & \textbf{@3} & \textbf{@5} & \\ \midrule
\rowcolor[HTML]{D3D3D3} 
\multicolumn{11}{c}{\textit{\textbf{In-Domain}}} \\ 
BM25* & 29.46 & 31.12 & 33.27 & 24.13 & 25.29 & 27.65 & 32.00 & 25.88 & 29.78 & 28.73 \\
EmbSim* & 63.67 & 61.03 & 65.37 & 49.11 & 42.27 & 46.56 & 53.00 & 46.40 & 52.73 & 53.35 \\
Re-Invoke* & 69.47 & - & 61.10 & 54.56 & - & 53.79 & 59.65 & - & 59.55 & 59.69 \\
IterFeedback & 71.64 & 71.29 & 76.31 & 62.65 & 55.58 & 60.62 & 73.85 & 64.06 & 69.02 & 67.22 \\
ToolGen & 69.47 & 72.26 & 79.12 & 46.77 & 53.58 & 62.45 & 77.06 & \underline{76.44} & \textbf{85.48} & 69.18 \\
ToolRetriever & \underline{81.91} & \underline{82.05} & \textbf{85.57} & \underline{75.92} & \underline{69.61} & \textbf{75.40} & \underline{79.82} & 72.42 & 77.11 & \underline{77.76} \\
\textbf{ToolOmni} & \textbf{86.07} & \textbf{84.49} & \underline{85.56} & \textbf{81.50} & \textbf{73.13} & \underline{74.11} & \textbf{84.40} & \textbf{76.81} & \underline{77.25} & \textbf{80.37} \\ \midrule
\rowcolor[HTML]{D3D3D3} 
\multicolumn{11}{c}{\textit{\textbf{Multi-Domain}}} \\ 
BM25* & 22.77 & 22.64 & 25.61 & 18.29 & 20.74 & 22.18 & 10.00 & 10.08 & 12.33 & 18.29 \\
EmbSim* & 54.00 & 50.82 & 55.86 & 40.84 & 36.67 & 39.55 & 18.00 & 17.77 & 20.70 & 37.13 \\
ToolGen & 68.84 & 72.00 & 78.76 & 46.77 & 53.55 & 62.43 & 75.68 & \textbf{75.26} & \textbf{84.48} & 68.64 \\
IterFeedback & 71.77 & 70.49 & 74.82 & 63.18 & 55.41 & 61.32 & 67.43 & 59.23 & 63.14 & 65.20 \\
ToolRetriever & \underline{81.03} & \underline{80.88} & \underline{84.52} & \underline{75.91} & \underline{69.53} & \textbf{75.21} & \underline{77.52} & 69.15 & \underline{74.24} & \underline{76.44} \\
\textbf{ToolOmni} & \textbf{86.07} & \textbf{84.09} & \textbf{84.80} & \textbf{80.63} & \textbf{72.96} & \underline{73.91} & \textbf{78.90} & \underline{71.44} & 71.82 & \textbf{78.29} \\ \bottomrule
\end{tabular}
\caption{Main results of tool retrieval performance (NDCG@$k$ \%) on ToolBench. \textbf{ToolOmni} achieves the best performance across most metrics. The best performance is boldfaced, while the second-best performance is underlined. \textbf{Note}: Methods marked with \textbf{*} report results cited from ToolGen~\cite{DBLP:conf/iclr/WangHJWB025}.}
\label{tab:retrieval_results}
\vspace{-5pt} 
\end{table*}


\section{Experiment}
In this section, we conduct extensive experiments on the ToolBench benchmark to answer the following Research Questions (RQs): 
\textbf{RQ1:} How does ToolOmni perform in retrieving relevant tools from massive, open-world repositories compared to existing baselines? 
\textbf{RQ2:} Can ToolOmni effectively execute complex, multi-step tasks in an end-to-end manner, surpassing pipeline and unified approaches? 
\textbf{RQ3:} Does the proposed framework demonstrate robustness against retrieval noise and generalization capabilities across unseen tools and domains? 
\textbf{RQ4:} What are the individual contributions of the core components (e.g., iterative retrieval, training stages) and hyperparameters to the overall performance?
\subsection{Experimental Setup}
We evaluate ToolOmni on \textbf{ToolBench} \cite{DBLP:conf/iclr/QinLYZYLLCTQZHT24,guo2024stabletoolbench}, testing across three difficulty levels (I1--I3) and three generalization splits (Instruction, Tool, Category) to assess robustness.
Retrieval performance is measured by \textbf{NDCG@$k$} ($k \in \{1, 3, 5\}$), while end-to-end execution is evaluated using \textbf{Solvable Pass Rate (SoPR)} and \textbf{Win Rate (SoWR)}, computed by a GPT-5 judge.
We initialize our model with \textbf{Qwen3-4B-Instruct}~\cite{yang2025qwen3} and train it via the proposed decoupled GRPO ($G=5, T=1.0$) on 8 NVIDIA H100 GPUs.
For comparison, we benchmark against competitive baselines across both tasks.
\textbf{For retrieval}, we consider sparse (BM25 \cite{robertson2009probabilistic}), dense (ToolRetriever \cite{DBLP:conf/iclr/QinLYZYLLCTQZHT24},EmbSim\footnote{OpenAI’s sentence embedding model:\texttt{text-embedding-3-large}}), and refinement methods (IterFeedback \cite{xu2024enhancing}, Re-Invoke \cite{DBLP:conf/emnlp/ChenYSWCBLP24}), alongside the generative ToolGen \cite{DBLP:conf/iclr/WangHJWB025}.
\textbf{For execution}, we evaluate pipeline agents (GPT, ToolLLaMA \cite{DBLP:conf/iclr/QinLYZYLLCTQZHT24}) paired with ToolRetriever, and the unified generative model ToolGen \cite{DBLP:conf/iclr/WangHJWB025}.
Further details are provided in \textbf{Appendix \ref{sec:appendix_setup}}.

\subsection{Tool Retrieval Performance(RQ1)}

We evaluate retrieval under two settings: 
\textit{In-Domain}, where the search space is restricted to the specific subset of tools relevant to the current test group; and 
\textit{Multi-Domain}, where the agent must identify the correct tools from the entire repository of over \textbf{16,000} APIs.
Regarding the mechanism, baselines typically employ static one-shot retrieval.
ToolGen uses generative retrieval, while IterFeedback performs multi-turn query refinement. 
Similarly, ToolOmni adopts an agentic iterative retrieval paradigm. To ensure a fair comparison, we align the search budget of ToolOmni with IterFeedback, limiting both to a maximum of 4 retrieval turns.

As shown in Table \ref{tab:retrieval_results}, \textbf{ToolOmni} achieves superior performance across the majority of metrics, particularly in the most rigorous Multi-Domain setting where it attains the highest average NDCG of \textbf{78.29\%}. It significantly outperforms strong baselines like ToolRetriever and ToolGen in top-1 (@1) and top-3 (@3) precision, demonstrating its superior ability to accurately pinpoint the "golden tools" from a massive repository.

Notably, ToolOmni occasionally scores slightly lower on \textbf{NDCG@5}. This is an expected side effect of our \textit{proactive selection mechanism}. Unlike standard retrievers that always return a fixed top-$k$ list, ToolOmni autonomously decides when to stop searching. If a task requires only 1 or 2 tools, our model outputs a concise set rather than padding it with irrelevant candidates to fill the top-5 slots. Thus, this performance gap reflects a preference for efficiency and precision over merely maximizing recall metrics.

\begin{table*}[t]
\centering
\small
\renewcommand\arraystretch{1.1}
\setlength{\tabcolsep}{8pt}
\begin{tabular}{l|cccc|cccc}
\toprule
\multirow{2}{*}{\textbf{Method}} & \multicolumn{4}{c|}{\textbf{Solvable Pass Rate (SoPR)}} & \multicolumn{4}{c}{\textbf{Solvable Win Rate (SoWR)}} \\
\cline{2-9}
 & \textbf{I1} & \textbf{I2} & \textbf{I3} & \textbf{Avg.} & \textbf{I1} & \textbf{I2} & \textbf{I3} & \textbf{Avg.} \\ \midrule

\rowcolor[HTML]{D3D3D3} 
\multicolumn{9}{c}{\textit{\textbf{With Ground-Truth Tools}}} \\ 
GPT-3.5 & \underline{56.60} & \underline{47.80} & \textbf{54.64} & \underline{53.01} & - & - & - & - \\
ToolLlama-v2 & 43.60 & 45.80 & 39.30 & 42.90 & \underline{38.65} & \underline{55.66} & \underline{32.79} & \underline{42.37} \\
ToolLlama-v1 & 24.20 & 29.20 & 12.30 & 21.90 & 28.22 & 42.45 & 16.39 & 29.02 \\
ToolGen (3B) & 40.80 & 43.40 & 21.30 & 35.17 & 39.26 & 47.17 & 14.75 & 33.73 \\
ToolGen (7B) & 44.20 & 42.90 & 32.00 & 39.70 & 35.58 & 45.28 & 16.39 & 32.42 \\
\textbf{ToolOmni} & \textbf{60.70} & \textbf{50.90} & \underline{52.50} & \textbf{54.70} & \textbf{52.76} & \textbf{56.60} & \textbf{40.98} & \textbf{50.11} \\ \midrule

\rowcolor[HTML]{D3D3D3} 
\multicolumn{9}{c}{\textit{\textbf{With Retriever (End-to-End)}}} \\ 
GPT-3.5 & \underline{51.43} & 41.19 & 34.43 & 42.35 & \underline{43.56} & 46.23 & \underline{29.51} & \underline{39.77} \\
ToolLlama-v2 & 43.10 & \underline{46.70} & \underline{40.20} & \underline{43.33} & 42.94 & \underline{50.00} & 14.75 & 35.90 \\
ToolLlama-v1 & 28.05 & 29.20 & 13.90 & 23.72 & 26.38 & 33.96 & 11.48 & 23.94 \\
ToolGen (3B) & 37.40 & 40.60 & 30.30 & 36.10 & 36.20 & 29.25 & 9.84 & 25.10 \\
ToolGen (7B) & 39.60 & 40.60 & 23.00 & 34.40 & 30.06 & 29.25 & 16.39 & 25.23 \\
\textbf{ToolOmni} & \textbf{59.20} & \textbf{58.10} & \textbf{45.10} & \textbf{54.13} $\uparrow$\textcolor{red}{\textbf{10.8}} & \textbf{50.31} & \textbf{57.55} & \textbf{42.62} & \textbf{50.16} $\uparrow$\textcolor{red}{\textbf{10.5}} \\ \bottomrule

\end{tabular}
\caption{End-to-end execution performance on ToolBench. We report both \textbf{Solvable Pass Rate (SoPR \%)} and \textbf{Solvable Win Rate (SoWR \%)}. ToolOmni demonstrates robust superiority, significantly outperforming GPT-3.5 and ToolLlama-v2 in end-to-end settings.}
\label{tab:execution_sopr_sowr}
\vspace{-5pt} 
\end{table*}

\subsection{Tool Execution Performance(RQ2)}
We assess execution performance in two scenarios: \textbf{(1)With Golden Truth} and \textbf{(2)With Retriever} (end-to-end setting).
For the end-to-end comparison, pipeline baselines are paired with ToolRetriever to perform a single initial search. ToolOmni and ToolGen utilize their internal retrieval mechanisms.
To ensure fair comparison during execution, we enforce a uniform interaction budget: the maximum number of tool execution turns is capped at 6 for all models.

Table \ref{tab:execution_sopr_sowr} details the execution results. 
With Golden Tools, \textbf{ToolOmni} attains the highest average SoPR of \textbf{54.70\%}, surpassing GPT-3.5 (53.01\%) and ToolLlama-v2 (42.90\%), validating its superior reasoning capabilities.
In the End-to-End setting, this advantage widens significantly. ToolOmni achieves \textbf{54.13\%} SoPR, outperforming the GPT-3.5 pipeline (42.35\%) by \textbf{+11.78\%} and doubling the gain over the unified model ToolGen (36.10\%). 
These results confirm the efficacy of our decoupled optimization strategy.

Furthermore, \textbf{ToolOmni} demonstrates robust superiority regarding the response quality metric \textbf{Solvable Win Rate (SoWR)}
In the end-to-end setting, it achieves an average SoWR of \textbf{50.16\%}, significantly surpassing the strongest open-source baseline ToolLlama-v2 (35.90\%) and the GPT-3.5 pipeline (39.77\%). 
This indicates that ToolOmni not only correctly solves more queries but also generates more precise and coherent answers.

\subsection{Robustness Analysis(RQ3)}
\begin{table}[t]
\centering
\small
\setlength{\tabcolsep}{6pt}

\begin{tabular}{l|ccc}
\toprule
\textbf{Method} & \textbf{Tool Gen.} & \textbf{Category Gen.} & \textbf{Average} \\ \midrule
GPT-3.5 & \underline{50.60} & \underline{42.10} & \underline{46.35} \\
ToolLlama-v2 & 39.20 & 39.40 & 39.30 \\
ToolGen (7B) & 33.20 & 40.10 & 36.65 \\
\textbf{ToolOmni} & \textbf{52.20} & \textbf{55.95} & \textbf{54.08} \\ \bottomrule
\end{tabular}
\caption{Generalization performance (SoPR \%) in the \textbf{End-to-End} setting. We report performance on Tool Gen. and Category Gen.. \textbf{ToolOmni} achieves the best generalization robustness.}
\label{tab:generalization_sopr}
\vspace{-10pt} 
\end{table}
\paragraph{Robustness Analysis on Generalization Splits.}
As presented in Table \ref{tab:generalization_sopr}, \textbf{ToolOmni} demonstrates exceptional robustness against distribution shifts. 
In the \textbf{Tool Generalization} setting (unseen tools within known categories), it achieves a SoPR of \textbf{52.20\%}, effectively adapting to new API signatures.
More impressively, in the rigorous \textbf{Category Generalization} setting (entirely novel domains), \textbf{ToolOmni} attains a remarkable score of \textbf{55.95\%}. 
This performance surpasses a strong pipeline baseline ChatGPT (42.10\%) by a substantial margin of \textbf{+13.85\%}.
This result indicates that while generative baselines like ToolGen tend to overfit to the specific tool seen during training, ToolOmni successfully learns the \textit{universal meta-skills} of tool usage—such as parameter inference and error recovery. 
Consequently, it can transfer its reasoning capabilities to completely new domains without requiring domain-specific fine-tuning.

\begin{figure}[t]
    \centering
    \begin{minipage}[c]{0.55\linewidth}
        \centering
        \includegraphics[width=\linewidth]{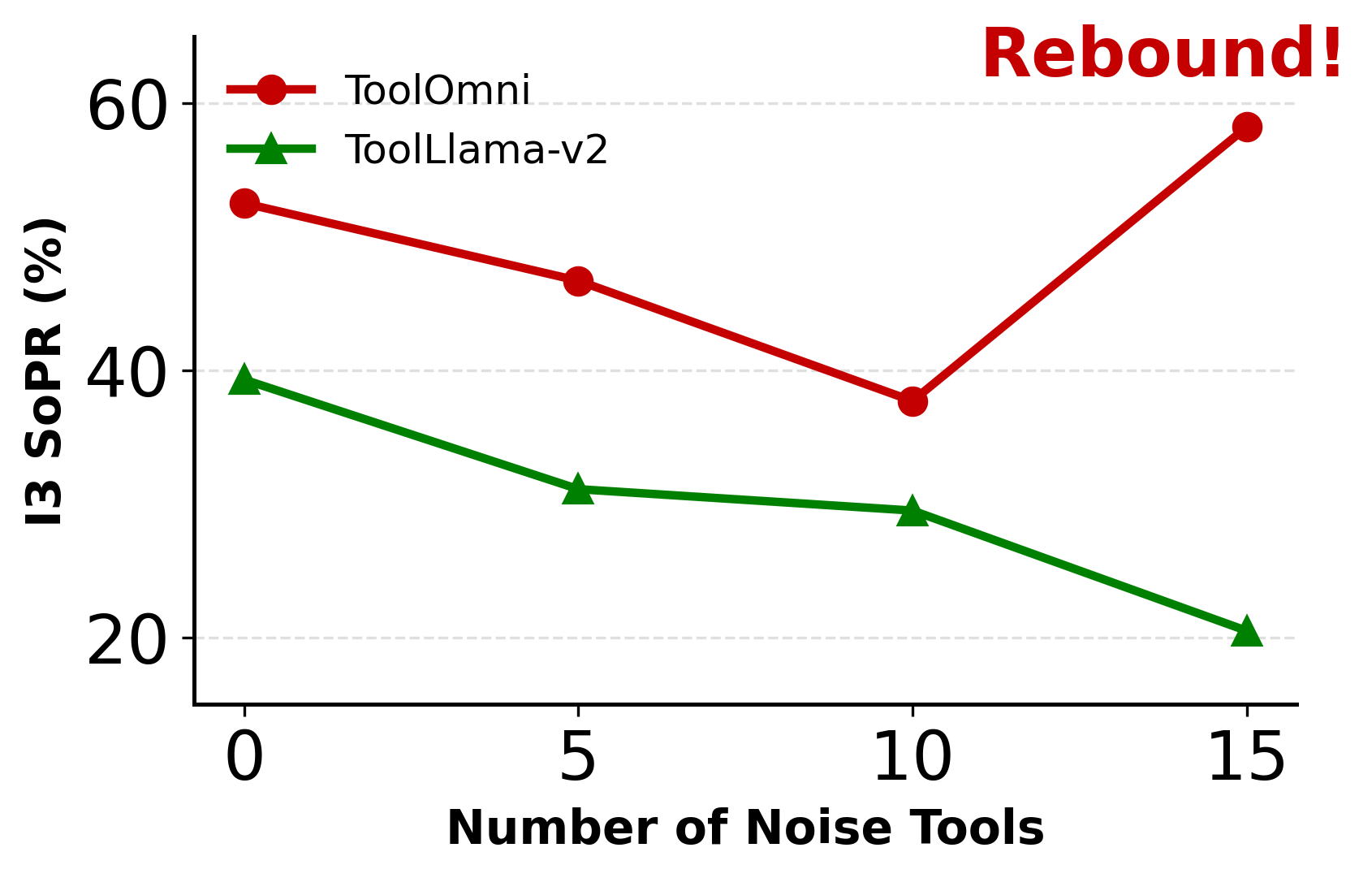}
        \caption{Robustness against adversarial noise on complex I3 tasks.}
        \label{fig:noise_robustness}
    \end{minipage}
    \hfill
    \begin{minipage}[c]{0.42\linewidth}
        \centering
        \small
        \makeatletter\def\@captype{table}\makeatother 
        \vspace{5pt} 

        \setlength{\tabcolsep}{6pt}
        \begin{tabular}{c|c}
            \toprule
            \textbf{Top-$k$} & \textbf{Avg. NDCG} \\ \midrule
            1 & 62.8 \\
            3 & 75.8 \\
            \textbf{5} & \textbf{78.3} \\
            7 & \underline{77.6} \\
            9 & 73.2 \\ \bottomrule
        \end{tabular}
        \caption{Ablation study on retrieval strategy. ``5'' is the default setting.}
        \label{tab:sensitivity_topk}
    \end{minipage}
\end{figure}

\paragraph{Robustness against Retrieval Noise.}
To test resilience, we injected adversarial tools that were retrieved by a dense model but excluding the ground truth—into the context at levels $N \in \{0, 5, 10, 15\}$.
As shown in Figure \ref{fig:noise_robustness}, \textbf{ToolLlama-v2} degrades monotonically (39.3\% $\to$ 20.5\%), revealing its susceptibility to semantic distraction.
In contrast, \textbf{ToolOmni} exhibits an adaptive resilience pattern. After an initial decline, its accuracy significantly \textbf{rebounds} to \textbf{58.2\%} at $N=15$.
This indicates that as the candidate pool expands, our agent adaptively identifies functionally similar \textit{alternative tools} within the adversarial set, showcasing a flexible reasoning capability that leverages noise to mitigate single-tool failures, transcending rigid ground-truth matching\cite{DBLP:conf/sigir/CuconasuTSFCMTS24}.

%
\subsection{Ablation Study(RQ4)}
To validate the contribution of our core components, we conduct comprehensive ablation studies across three critical dimensions: the multi-stage training pipeline, the iterative retrieval mechanism, and the specific reinforcement learning designs. For the macroscopic architecture, we first compare ToolOmni against three variants: (1) \textbf{w/o Iterative Retrieval}, a static baseline restricted to single-round retrieval; (2) \textbf{w/o RL}, the model trained solely via SFT; and (3) \textbf{w/o SFT}, the model trained directly via RL from the base LLM, skipping the cold-start phase. Furthermore, to disentangle the effectiveness of our Decoupled Multi-Objective GRPO algorithm, we introduce specific micro-ablations for the RL components, including \textbf{w/o Filter}, \textbf{Combined Update}, and \textbf{Vanilla GRPO}.

\begin{figure}[t] 
    \centering
    \begin{minipage}[c]{0.48\linewidth} 
        \centering
        \small 
        \makeatletter\def\@captype{table}\makeatother 
        \setlength{\tabcolsep}{3pt} 
        \begin{tabular}{l|c|c}
            \toprule
            \multirow{2}{*}{\textbf{Settings}} & \textbf{Ret.} & \textbf{Exec.} \\
             & \textbf{NDCG} & \textbf{SoPR} \\ \midrule
            \textbf{Full} & \textbf{78.3} & \textbf{52.5} \\ 
            w/o SFT & \underline{77.8} & \underline{43.4} \\ 
            w/o RL & 49.3 & 23.8 \\ \bottomrule
        \end{tabular}
        \caption{Ablation on training stages. \textbf{Full ToolOmni} integrated with SFT as well as RL achieves the optimal results.}
        \label{tab:ablation_sft_rl}
    \end{minipage}
    \hfill 
    \begin{minipage}[c]{0.48\linewidth}
        \centering
        \includegraphics[width=\linewidth]{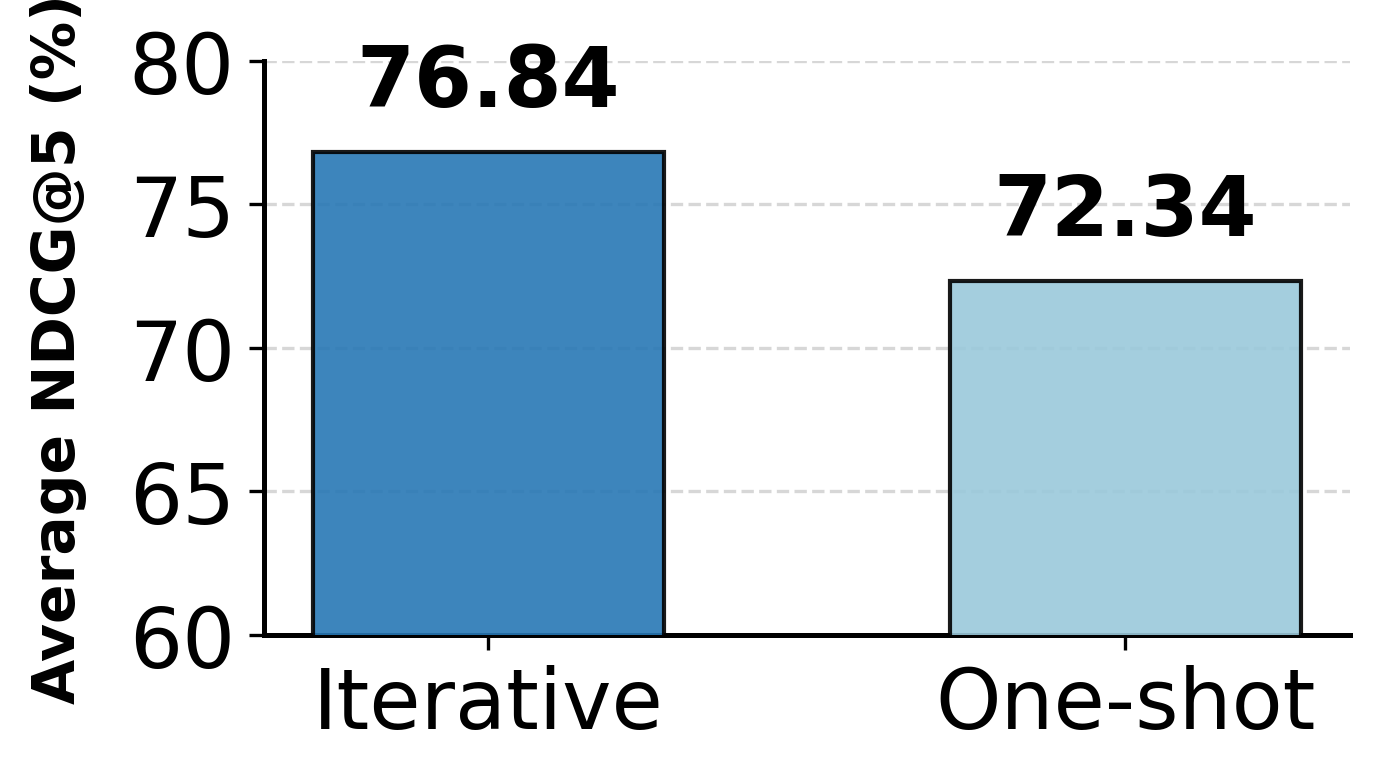}
        \makeatletter\def\@captype{figure}\makeatother 
        \caption{Ablation on retrieval strategy. \textbf{Iterative Retrieval} boosts NDCG@5 over the One-shot baseline.}
        \label{fig:ablation_retrieval}
    \end{minipage}
\end{figure}
\paragraph{Impact of Training Stages.}
Table \ref{tab:ablation_sft_rl} disentangles the impact of training stages. For \textbf{Retrieval}, the removal of RL causes a drastic drop in NDCG (-28.99\%), while removing SFT has a negligible impact (-0.51\%), indicating that retrieval quality is primarily driven by RL-based exploration.
Conversely, for \textbf{Execution}, the \textit{w/o SFT} variant suffers a significant performance gap compared to the full model (43.40\% vs. 52.50\%). This confirms that SFT is essential as a cold-start mechanism, providing the necessary reasoning foundation that enables RL to optimize for complex tool-use scenarios effectively.



\paragraph{Impact of Iterative Retrieval.}
We further analyze the effectiveness of the retrieval strategy by comparing the Average NDCG@5 across all splits. As shown in Figure \ref{fig:ablation_retrieval}, the \textbf{Iterative Retrieval} mechanism employed by ToolOmni achieves an average score of \textbf{76.84\%}, surpassing the static \textbf{One-shot Retrieval} baseline by \textbf{+4.5\%}. 
This indicates that the ability to iteratively refine search queries allows the agent to dynamically adjust its search intent and filter out noise, effectively pinpointing the most utility-oriented tools required for execution.

\paragraph{Impact of RL Components Design.} 
To validate the specific contributions of our proposed Decoupled Multi-Objective GRPO algorithm, we conduct a detailed ablation study on its key components.

\begin{figure}[ht]
    \centering
    \includegraphics[width=0.85\linewidth]{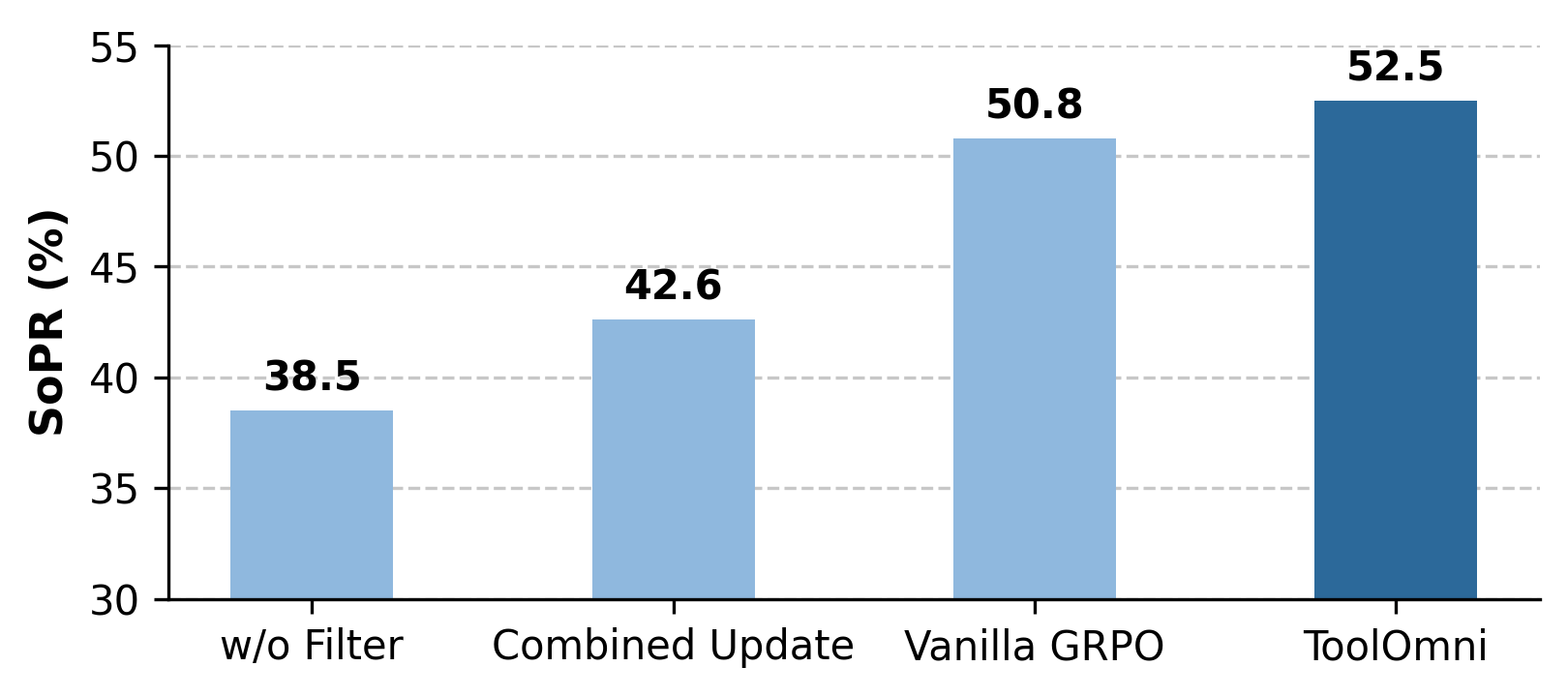} 
    \caption{Ablation study of RL components on the ToolBench. Decoupled optimization and trajectory filtering are crucial for execution performance.}
    \label{fig:rl_ablation}
\vspace{-5pt}
\end{figure}

As illustrated in Figure~\ref{fig:rl_ablation}, our full \textbf{ToolOmni} model achieves the highest SoPR of 52.5\%. The most substantial performance drop occurs when removing the trajectory filtering mechanism, falling to 38.5\%, confirming that forcing the execution model to train on invalid contexts severely hinders its ability to reason and answer questions correctly. Furthermore, the \textbf{Combined Update} variant suffers a notable decline (42.6\%), highlighting that directly summing gradients from the retrieval and execution phases leads to destructive gradient conflicts. Finally, replacing our decoupled reward design with \textbf{Vanilla GRPO} degrades performance (50.8\%), demonstrating that assigning fine-grained, independent rewards provides more precise supervision than a sparse, unified reward.
\subsection{Hyperparameter Sensitivity(RQ4)}
\paragraph{Sensitivity to Retrieval Count ($k$).}
We investigate how the number of retrieved tools provided to the agent affects overall performance. Table \ref{tab:sensitivity_topk} reports the Average NDCG across different retrieval counts $k \in \{1, 3, 5, 7, 9\}$.

The performance follows a clear \textbf{inverted U-shaped trajectory}. At low values ($k=1$), the model suffers significantly (62.84\%), as the constrained context often fails to include the necessary ground-truth tools (low recall). As $k$ increases to 5, performance peaks at \textbf{78.29\%}, indicating an optimal balance where the context is rich enough to cover required functionalities without being overwhelming.
However, increasing $k$ further to 9 leads to a noticeable decline (73.16\%). 
This degradation suggests that an excessively long context introduces irrelevant "noise tools," which dilutes the agent's attention and complicates the reasoning process. 
Thus, our default setting of $k=5$ proves to be the most robust configuration for open-world scenarios.

\paragraph{Sensitivity to Format Reward Weight.}
We further investigate the impact of the format reward weight on execution performance.
Specifically, We varied the format weight across $\{0.2, 0.4, 0.6, 0.8\}$. The results demonstrate a clear degradation trend when the format constraint becomes too dominant. 
While a moderate weight of 0.4 yields a peak SoPR of 44.3\%, further increasing the weight to 0.6 and 0.8 causes the SoPR to drop significantly to 41.0\% and 38.5\%, respectively. 
This indicates that an excessively high format reward forces the model to overly prioritize strict syntactic compliance at the expense of exploring complex reasoning paths and functional tool chains. 
Consequently, a lower format weight strikes the optimal balance, ensuring structural validity without stifling the agent's problem-solving capabilities in intricate open-world scenarios.
\begin{figure}[ht]
    \centering
    \includegraphics[width=0.85\linewidth]{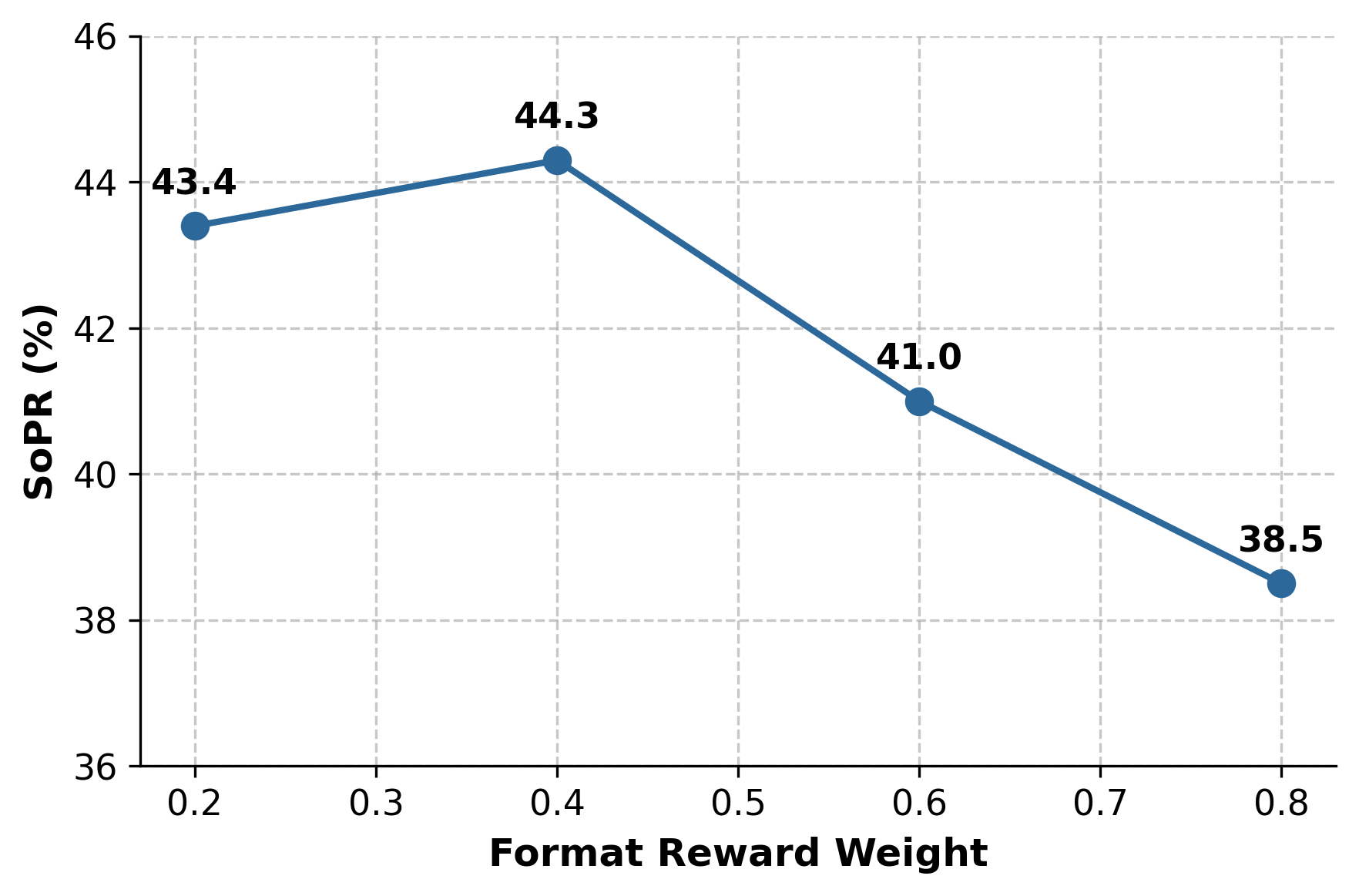} 
    \caption{Sensitivity analysis of the format reward weight on the ToolBench.}
    \label{fig:format_weight}
\vspace{-5pt}
\end{figure}
\subsection{Computational Efficiency}
\label{sec:efficiency}
A potential concern regarding the iterative proactive retrieval mechanism is the introduction of additional computational overhead. To address this, we analyze the overall efficiency by comparing the average number of search queries and actual tool execution calls per instruction across different models on the ToolBench benchmark.
\begin{table}[ht]
\centering
\small
\setlength{\tabcolsep}{4pt}
\begin{tabular}{l|ccccc}
\toprule
\textbf{Metric} & \textbf{ToolLlama} & \textbf{ToolGen} & \textbf{ChatGPT} & \textbf{ToolOmni} \\
\midrule
Search    & 1.00 & -    & 1.00 & 3.02 \\
Tool & 2.40 & 2.75 & 3.98 & 2.47 \\
\bottomrule
\end{tabular}
\caption{Average number of search times and tool execution calls per query on the ToolBench Benchmark.}
\label{tab:efficiency}
\vspace{-10pt}
\end{table}

As shown in Table~\ref{tab:efficiency}, ToolOmni performs more search operations (3.02) than static baselines (1.00). However, this proactive exploration effectively filters noise, resulting in significantly fewer actual tool invocations (2.47 vs. ChatGPT's 3.98). In real-world deployments, external API calls are the primary system bottleneck due to network latency and strict rate limits, whereas local embedding retrieval is computationally inexpensive. By trading a slight increase in cheap retrievals for a substantial reduction in costly, error-prone API calls, ToolOmni optimizes overall computational efficiency without sacrificing task success.

\section{Conclusion}
In this paper, we presented \textbf{ToolOmni}, a comprehensive agentic framework that enables open-world tool use via agentic learning with proactive retrieval and grounded execution. 
To address the inherent challenges of sparse rewards and sequential dependencies in this domain, we proposed a novel \textbf{Decoupled Multi-Objective GRPO} algorithm. 
Extensive experiments on the ToolBench benchmark demonstrate that \textbf{ToolOmni} achieves state-of-the-art performance in retrieval and end-to-end generation.
Notably, our agent exhibits exceptional robustness against retrieval noise and strong generalization capabilities across unseen tools and domains.
These results validate that equipping LLMs with the ability to iteratively reason about tool selection and execution is the key to scalable and robust tool learning. 
Future work will explore extending this decoupled framework to multi-modal tools and more complex, long-horizon planning tasks.

\clearpage
\section*{Limitations}
While ToolOmni demonstrates strong generalization, we acknowledge two primary limitations. First, our architecture relies on a rigid cascaded paradigm. While stable, this separation restricts the agent's flexibility in extremely complex tasks that require on-the-fly tool chain discovery based on intermediate execution results. Second, due to computational constraints, ToolOmni is currently trained exclusively on the Qwen3-4B base model. Consequently, the upper bound of its performance and the potential emergent reasoning capabilities from scaling our framework to larger foundation models remain unexplored.

\section*{Acknowledgements}
This work is jointly supported by National Natural Science Foundation of China (Grant No.62422603), the Guangdong Basic and Applied Basic Research Foundation (Grant No.2024B0101050003), and Shenzhen Science and Technology Program (Grant No.ZDSYS20230626091203008). We sincerely thank all anonymous reviewers and Area Chairs for their detailed and careful reviews and valuable suggestions, which have significantly improved our work.

\bibliography{custom}

\appendix

\section{Data Curation}
\label{sec:appendix_data}
\subsection{ToolBench}
\label{sec:appendix_toolbench}
ToolBench \cite{DBLP:conf/iclr/QinLYZYLLCTQZHT24} serves as a pioneering open-source benchmark for tool learning, constructed by scraping real-world APIs from RapidAPI. It encompasses a massive repository of \textbf{16,464 APIs} distributed across \textbf{49 distinct categories} (e.g., Social Media, E-commerce, Weather), reflecting a true open-world environment. 
Each API is documented with a JSON-based schema specifying its name, description, and calling parameters.

To generate high-quality instruction-tuning data, ToolBench utilizes a \textbf{Depth-First Search Decision Tree(DFSDT)} mechanism powered by ChatGPT. This process efficiently explores the vast action space to synthesize diverse and valid solution paths, resulting in over 126k tool-use instructions.

Crucially, these instructions cover a wide spectrum of complexity, ranging from single-tool tasks (I1) to intricate scenarios requiring the composition of multiple tools from the same category (I2) or across different collections (I3). 
This hierarchical diversity makes it an ideal testbed for evaluating an agent's generalization and reasoning capabilities.

\subsection{Source and Difficulty Stratification}
Our dataset is derived from the \textbf{ToolBench} corpus.
To construct a balanced and challenging dataset, we first assess the difficulty of each query. We employ \textbf{ToolRetriever} to perform a preliminary dense retrieval, returning the top-5 tool candidates for each user query. An instance is classified based on recall success:
\begin{itemize}
    \item \textbf{Easy Data}: All ground-truth "golden tools" are successfully retrieved within the top-5 candidates.
    \item \textbf{Hard Data}: At least one golden tool is missed by the retriever.
\end{itemize}
We then perform stratified sampling, selecting \textbf{60,000 hard instances} and \textbf{20,000 easy instances}, totaling \textbf{80,000 queries}. This set serves as the foundational pool for our subsequent processing and is also utilized as the prompt source for the Reinforcement Learning (RL) stage.

\subsection{Retrieval Training Data}
From the initial 80,000 queries, we train a specialized retrieval-only model to generate candidate search trajectories. We then apply \textbf{Rejection Sampling} on these generations, filtering out trajectories where the model fails to locate the correct tools. This rigorous filtering yields a high-quality corpus of approximately \textbf{28,000 retrieval trajectories}, which is used for the retrieval component of the SFT stage.

\subsection{Execution Training Data}
For the execution phase, we process the same pool of 80,000 instances to align with our agentic format. Since original ToolBench data lacks explicit reasoning steps, we employ \textbf{Qwen-2.5-72B-Instruct} to augment the data by generating detailed reasoning paths before each tool invocation.
To ensure correctness, we use \textbf{Qwen-2.5-32B} as an automated judge to verify the execution results. 
To improve the model's discrimination ability during training, we construct a mixed dataset comprising:
\begin{itemize}
    \item \textbf{Positive Samples (70\%)}: Trajectories verified as correct.
    \item \textbf{Negative Samples (30\%)}: Trajectories containing errors (e.g., hallucinated parameters or wrong tool selection).
\end{itemize}
This process results in a final execution dataset of approximately \textbf{33,000 trajectories}.

\subsection{Final Dataset Composition}
The \textbf{SFT Dataset} is the union of the 28,000 retrieval trajectories and 33,000 execution trajectories. The \textbf{RL Dataset} utilizes the initial pool of 80,000 queries as prompts to drive the online exploration and optimization process.

\section{Experiment Setup}
\label{sec:appendix_setup}
\subsection{Datasets and Metrics}
Our experiments are conducted on ToolBench \cite{DBLP:conf/iclr/QinLYZYLLCTQZHT24,guo2024stabletoolbench}, a comprehensive real-world tool benchmark containing 16000+ real-word apis. 
Following the split of \cite{DBLP:conf/iclr/QinLYZYLLCTQZHT24}, the test queries are categorized into three complexity levels: \textbf{I1} (single-tool instructions), \textbf{I2} (intra-category multi-tool instructions), and \textbf{I3} (intra-collection multi-tool instructions). 
To rigorously evaluate robustness, these scenarios are further stratified into three generalization splits: \textbf{Instruction (Inst)} generalization (unseen queries with seen tools), \textbf{Tool} generalization (unseen tools within known categories), and \textbf{Category (Cate)} generalization (unseen tools from entirely new domains).

We evaluate tool retrieval performance with \textbf{NDCG@k} (with $k \in \{1, 3, 5\}$).
For end-to-end evaluation, we report two key metrics: \textbf{Solvable Pass Rate (SoPR)}, which measures the percentage of queries successfully solved by the agent, and \textbf{Solvable Win Rate (SoWR)}, which indicates the proportion of answers that outperform those generated by a reference baseline. 
Both metrics are computed using GPT-5 Mini as an automated judge to ensure consistent and scalable evaluation.
\subsubsection{Retrieval Test Dataset}
The retrieval evaluation is conducted on the official ToolBench test set. Table \ref{tab:retrieval_stats} details the distribution of test queries across the three instruction complexity levels.

\begin{table}[h]
\centering
\small
\setlength{\tabcolsep}{15pt} 
\begin{tabular}{l|c}
\toprule
\textbf{Complexity Level} & \textbf{Number of Queries} \\ 
\midrule 
I1 (Single-tool)          & 796                        \\ 
I2 (Intra-category)       & 573                        \\ 
I3 (Intra-collection)     & 218                        \\ 
\midrule
\textbf{Total}            & \textbf{1587}              \\ 
\bottomrule
\end{tabular}
\caption{Statistics of the \textbf{ToolBench Retrieval Test Set}. The dataset is categorized by instruction complexity (I1: Single-tool, I2: Intra-category, I3: Intra-collection).}
\label{tab:retrieval_stats}
\end{table}

\subsubsection{Execution Test Dataset}
For the end-to-end execution evaluation, we utilize the curated \textbf{Solvable Test Queries} from StableToolBench to ensure robust assessment. The test set comprises six distinct generalization subsets, covering different levels of difficulty (G1, G2, G3) and generalization types (Instruction, Category, Tool). Table \ref{tab:execution_stats} presents the detailed statistics.
\begin{table}[h]
\centering
\small
\setlength{\tabcolsep}{10pt} 
\begin{tabular}{ll|c}
\toprule
\textbf{Difficulty} & \textbf{Subset Name} & \textbf{Count} \\ 
\midrule 
\multirow{3}{*}{\textbf{G1}} & Instruction Gen. (I1) & 163 \\ 
                            & Tool Gen. (I1)        & 158 \\ 
                            & Category Gen. (I1)    & 153 \\ 
\midrule
\multirow{2}{*}{\textbf{G2}} & Instruction Gen. (I2) & 106 \\ 
                            & Category Gen. (I2)    & 124 \\ 
\midrule
\textbf{G3}                 & Instruction Gen. (I3) & 61  \\ 
\midrule
\multicolumn{2}{l|}{\textbf{Total}} & \textbf{765} \\ 
\bottomrule
\end{tabular}
\caption{Statistics of the \textbf{StableToolBench Solvable Test Set}. Queries are stratified by generalization difficulty (G1, G2, G3) and type.}
\label{tab:execution_stats}
\end{table}

\subsubsection{Evaluation Reliability}
To address potential biases and stochastic instability associated with utilizing LLM-as-a-Judge for execution metrics, we conduct a multi-faceted reliability analysis to ensure the objectivity and consistency of our automated evaluation \cite{zheng2023judging}.
\paragraph{Human-Model Alignment.}
We randomly sampled a subset of evaluation set and conducted a comprehensive human evaluation. Three human annotators scored the trajectories using a three-point scheme: -1 (unsolved), 0 (unsure), and 1 (solved). 
We then calculated the Kendall tau correlation coefficient \cite{kendall1938new} between the GPT-5 judgments and human annotations.
The correlation result reached \textbf{0.847}, indicating a strong consistency between our automated evaluation and human assessment, which validates the reliability of our main experimental results.
\paragraph{Cross-Model Consistency.}
To further mitigate potential model-selection bias and ensure that the performance gains of ToolOmni are not backbone-specific, 
we evaluated our approach using three different strong models as judges: GPT-5, Gemini-3, and Qwen2.5-32B. 
As shown in Table~\ref{tab:multi_judge}, ToolOmni consistently achieves the state-of-the-art overall performance across all judge models. 
While the absolute SoPR values vary across different judges (e.g., Qwen tends to give higher scores overall), the relative ranking of the methods remains stable, further confirming the robustness of our framework.

\begin{table}[ht]
\centering
\small
\begin{tabular}{l|ccc|c}
\toprule
\textbf{Method} & \textbf{GPT-5} & \textbf{Gemini-3} & \textbf{Qwen-32B} & \textbf{Mean} \\
\midrule
ChatGPT    & 41.68 & 32.18 & 54.60 & 42.82 \\
Llama-v2   & 42.22 & 30.28 & 53.00 & 41.83 \\
Llama-v1   & 28.47 & 21.92 & 46.55 & 32.31 \\
ToolGen-3B & 35.62 & 21.68 & 36.62 & 31.31 \\
ToolGen-7B & 34.73 & 15.62 & 32.83 & 27.73 \\
ToolOmni   & 54.08 & 31.95 & 79.15 & \textbf{55.06} \\
\midrule
\textbf{Average} & 39.47 & 25.61 & 50.46 & - \\
\bottomrule
\end{tabular}
\caption{Average SoPR (\%) on ToolBench judged by different LLM backbones.}
\label{tab:multi_judge}
\end{table}

\subsection{Comparison Baselines.}
To validate the effectiveness of ToolOmni, we benchmark it against a comprehensive set of competitive methods. 
For tool retrieval, we compare with traditional sparse retrievers like BM25 \cite{robertson2009probabilistic} and dense retrievers such as ToolRetriever \cite{DBLP:conf/iclr/QinLYZYLLCTQZHT24}. 
Additionally, we evaluate advanced query refinement strategies, including Re-Invoke \cite{DBLP:conf/emnlp/ChenYSWCBLP24} and IterFeedback \cite{xu2024enhancing}, as well as ToolGen \cite{DBLP:conf/iclr/WangHJWB025}, which adopts a generative paradigm for retrieval. 
For tool execution, our baselines encompass both proprietary and open-source models: we employ  ChatGPT as strong zero-shot references, and compare against ToolLLaMA \cite{DBLP:conf/iclr/QinLYZYLLCTQZHT24}, the state-of-the-art SFT baseline on ToolBench. 
We also include ToolGen \cite{DBLP:conf/iclr/WangHJWB025} again to assess the performance of unified generative frameworks in end-to-end scenarios.

\subsection{Implementation Details.} 
We initialize ToolOmni upon the Qwen3-4B-Instruct \cite{yang2025qwen3}. 
Regarding the reward configuration, we set the format weight to $0.2$ and the performance weight to $0.8$ for both phases (i.e., $\alpha_1=0.2, \alpha_{2}=0.8$ for retrieval; $\beta_1=0.2, \beta_2=0.8$ for execution)
For the reinforcement learning stage, we configure the GRPO group size to $G=5$ and set the sampling temperature to 1.0 to encourage exploration during training.
The retrieval module employs ToolRetriever as the dense embedding model. For the execution stage, we implement a hybrid environment that integrates real-world tools with MirrorAPI \cite{guo2024stabletoolbench}; this configuration ensures both the authenticity of tool interactions and the robustness 
All models are trained for a single epoch on 8 NVIDIA H100 GPUs.

\subsection{Execution Environment}

To ensure both the authenticity of tool interactions and the robustness of the evaluation, we implement a hybrid execution environment that synergizes real-world API response logs with the MirrorAPI simulator \cite{guo2024stabletoolbench}. 

Specifically, the environment operates on a fallback-based retrieval logic: upon a tool invocation, the system first queries a comprehensive repository of historical real-world API call records. If a matching execution trace is found, the authentic response is returned directly to ensure maximum fidelity. In cases where real-world records are unavailable, the system seamlessly transitions to MirrorAPI, which serves as a high-fidelity proxy. 
Crucially, MirrorAPI is not merely a deterministic mapping; it is pre-trained on an extensive corpus of real-world API interactions, allowing it to faithfully replicate the stochasticity and error distributions inherent in practical environments. This enables the simulation of various failure modes, such as missing parameters, authentication errors, and service timeouts, thereby providing a rigorous testbed for the model's error recovery capabilities.
To illustrate the model's interaction with this dynamic environment, we present a representative reasoning trajectory of ToolOmni in Box~\ref{box:trace}. 
\begin{tcolorbox}[
    colback=white, colframe=gray,
    title={ToolOmni Reasoning Trajectory under Execution Failures},
    fonttitle=\bfseries,
    label={box:trace}
]
\small
\textbf{1. Initial Tool Call (Failure)}

\tool{\{"category": "Advertising", "tool\_name": "Reqres", "api\_name": "Users", "tool\_input": \{\}\}}

\textbf{2. Environment Feedback (Simulated Error)}
\info{\{"error": "Missing input parameters.", "response": ""\}}

\textbf{3. Model Reasoning \& Adaptive Adjustment}
\reason{The initial call to retrieve user lists failed due to missing parameters. I will now switch to the "User by id" tool, providing the specific ID '1' to recover from this error.}

\textbf{4. Refined Tool Call (Success)}

\tool{\{"category": "Advertising", "tool\_name": "Reqres", "api\_name": "User by id", "tool\_input": \{"id": "1"\}\}}

\textbf{5. Final Environment Feedback}

\info{\{"error": "", "response": "\{"user\_id": 1, "name": "John Doe", "email": "john.doe@example.com"\}"\}}

\end{tcolorbox}
This execution flow confirms that our hybrid environment not only provides realistic feedback but also effectively evaluates the model's ability to detect, reason about, and recover from execution failures in complex, real-world scenarios.
\section{Case Study}
To provide a qualitative understanding of ToolOmni's superiority, we present comprehensive case studies across both Open-Domain and Oracle settings.
In the \textbf{Open-Domain scenario} (Fig.\ref{fig:case_studies_open_domain}), we demonstrate how ToolOmni's proactive iterative retrieval effectively filters out noise and locates critical tools that pipeline baselines often miss, thereby preventing downstream hallucinations.
In the \textbf{With Golden Truth setting} (Fig.\ref{fig:case_study_adaptive}), we highlight the agent's robust reasoning capabilities. Specifically, Case 2 illustrates how ToolOmni autonomously diagnoses API errors (e.g., missing tokens) and strategically pivots to alternative tools.
Together, these cases validate that ToolOmni is not merely a tool invoker, but a resilient problem solver capable of navigating the complexities of open-world environments.
\begin{figure*}[t]
\centering
\begin{tcolorbox}[colback=white, colframe=gray, title=\textbf{Case 1: Open-Domain Scenario}, width=0.98\textwidth]
    \footnotesize 
    \textbf{\faUser\ Instruction:} I want to explore different genres of movies and series available on streaming platforms. Can you provide me with the basic details and streaming availability of a movie or series with the keyword 'zombie' and released between the years 2000 and 2025? Also, fetch the details of the actors and crew associated with the title with Watchmode ID 3173903.
    \tcblower
    
    \begin{minipage}[t]{0.48\linewidth}
        \footnotesize
        \textbf{\faTimes\ Baseline (ToolLlama + ToolRetriever):}
        \vspace{2pt}
        
        \textit{[ToolRetriever Retrieval Result (Top-5)]} \\
        \begin{itemize}
            \item \textcolor{gray}{[ID:781] OTT - Search} \textcolor{red}{(Noise)}
            \item \textcolor{mygreen}{\textbf{[ID:782] OTT - Title Details}}
            \item \textcolor{mygreen}{\textbf{[ID:4580] Watchmode - Cast \& Crew}}
            \item \textcolor{gray}{[ID:778] OTT - Advanced Search} \textcolor{red}{(Wrong Tool for Streaming)}
            \item \textcolor{gray}{[ID:13631] Watchmode - Person} \textcolor{red}{(Noise)}
        \end{itemize}
        \textcolor{red}{\textit{Missing Golden Tool:}} \textbf{[ID:7585] Streaming Availability - Get Basic}
        
        \vspace{3pt}
        \textit{[Execution Phase]} \\
        \textcolor{red}{\texttt{Action 1:}} \texttt{OTT-Search(title="zombie", page=1)} \\
        \textcolor{red}{\textit{Error:}} Used basic search; returned random titles like "Zombie Outbreak" without checking year/genre. \\
        
        \textcolor{red}{\texttt{Action 2:}} \texttt{TitleDetails(imdbid="tt13411672")} \\
        \textit{Result:} Streaming information. \\
        
        \textcolor{red}{\texttt{Action 3:}} \texttt{TitleDetails(id="3173903")} \\
        \textcolor{red}{\textit{Error:}} Wrong tool! Used `TitleDetails` instead of `CastCrew` for Watchmode ID. Returned unrelated anime "Jujutsu Kaisen".
        
        \vspace{2pt}
        \textcolor{myred}{\textbf{Outcome: Hallucination.}} Falsely claims "Available on Netflix" and provides wrong cast info.
    \end{minipage}
    \hfill
    \vrule width 0.5pt
    \hfill
    \begin{minipage}[t]{0.48\linewidth}
        \footnotesize 
        \textbf{\faRobot\ ToolOmni (Ours):}
        \vspace{2pt}
        
        \textit{[Phase 1: Proactive Retrieval]} \\
        \search{fetch basic details and streaming availability of a movie or series} \\
        \info{\textcolor{mygreen}{\textbf{[ID:782] OTT - Title Details}} \textit{Get basic information \dots}}
        
        \search{confirm the keyword 'zombie' and released between the years 2000 and 2025, specifically for streaming platforms.} \\
        \info{\textcolor{mygreen}{\textbf{[ID:7585] Streaming Availability - Get Basic}} \textit{Get details of a show \dots}} \\
        \search{associated with Watchmode ID 3173903, specifically for streaming platforms and crew details.} \\
        \info{\textcolor{mygreen}{\textbf{[ID:4580] Watchmode - Title Cast \& Crew}} \textit{Return all people associated with the title \dots}} \\
        \finaltools{\textcolor{mygreen}{\textbf{ID: [782,7585,4580]}}} \\
        \vspace{2pt}
        \textit{[Phase 2: Grounded Execution]} \\
        \reason{To fulfill this request, I need to:1. Search for movies or series with the keyword 'zombie' and released between 2000 and 2025 using the OTT details search tool with a filter for release year.2. Fetch the basic details and streaming availability of such a movie or series using the Streaming Availability tool with a country parameter (e.g., US).3. Retrieve the cast and crew details for the title with Watchmode ID 3173903 using the Watchmode tool.} \\
        \tool{\{"tool": "OTT details", "api": "Search", "input": \{"title": "zombie", "start\_year": 2000, "end\_year": 2025\}\}} \\
        \info{\textit{Found "Zombie (2000)", "Zombie Nightmare"...}} \\
        \reason{Check streaming availability.} \\
        \tool{\{"tool": "Streaming Availability", "api": "Get Basic", "input": \{"country":"us", "imdb\_id": "tt0238784"\}\}} \\
        \info{\textit{Streaming Not Available}} \\
        \reason{Get crew details}
        \tool{\{"tool": "Watchmode", "api": "Title Cast \& Crew", "input": \{"id": "3173903"\}\}}
        \info{\textit{Crew details \dots}}
        \reason{The search for movies or series with the keyword 'zombie' and released between 2000 and 2025 returned valid results, including both movies and TV movies. Additionally, the cast and crew details for the title with Watchmode ID 3173903 have been successfully retrieved. Although streaming availability was not found in the US, the user's request is satisfied with the exploration of genres and the cast/crew details for the specified title.} \\
        \answer{Provided movie list (2000-2025), streaming status and correct cast list.} \\
        \textcolor{mygreen}{\textbf{Outcome: Solved}}
    \end{minipage}
\end{tcolorbox}

\caption{Qualitative comparison (Case Study). \textbf{Case 1} illustrates how pipeline baselines fail due to retrieval noise (selecting basic Search instead of Advanced Search) and missing tools (Streaming), leading to hallucinations. In contrast, \textbf{ToolOmni}'s proactive iterative retrieval precisely locates the golden toolset, enabling correct execution and grounded response generation. \textit{Note: The iterative process shown is condensed for brevity.}}

\label{fig:case_studies_open_domain}
\end{figure*}

\begin{figure*}[t]
\centering
\begin{tcolorbox}[colback=white, colframe=gray, title=\textbf{Case 2: Complex Reasoning With Golden Tools}, width=0.98\textwidth]
    \small
    \textbf{\faUser\ Instruction:} I'm a movie enthusiast and I'm interested in watching critically acclaimed films. Can you suggest some award-winning movies with high IMDb ratings? Also, provide the streaming platforms where these movies are available in the US.
    \tcblower
    
    \begin{minipage}[t]{0.48\linewidth}
        \textbf{\faTimes\ Baseline (ToolLlama):}
        \vspace{2pt}
        
        \textit{[Step 1]} \\
        \texttt{Action:} \texttt{Streaming.Search(country="US", services="netflix,prime...")} \\
        \textcolor{red}{\textit{Error:}} \texttt{Invalid country code.}
        
        \vspace{2pt}
        \textit{[Step 2]} \\
        \texttt{Action:} \texttt{Streaming.Search(country="US", ...)} \textit{(Retrying same args)} \\
        \textcolor{red}{\textit{Error:}} \texttt{Invalid country code.}
        
        \vspace{2pt}
        \textit{[Step 3-5]} \\
        \textcolor{gray}{(Repeatedly calls the same tool with minor variations, ignoring the error message.)}
        
        \vspace{4pt}
        \textcolor{red}{\textbf{Outcome: Give Up.}} The model entered a repetitive loop and failed to adjust its strategy, eventually terminating with \texttt{give\_up\_and\_restart}.
    \end{minipage}
    \hfill
    \vrule width 0.5pt
    \hfill
    \begin{minipage}[t]{0.48\linewidth}
        \textbf{\faRobot\ ToolOmni (Ours):}
        \vspace{2pt}
        
        \reason{The user wants to watch critically acclaimed films with high IMDb ratings and wants to know the streaming platforms where these movies are available in the US. To fulfill this request, I should start by searching for movies that are both critically acclaimed and have high IMDb ratings.} \\
        \tool{\{"tool": "Streaming Availability", "api": "Search Ultra", "input": \{"country": "us", "min\_imdb\_rating": 80\}\}} \\
        \info{Empty}
        
        \vspace{2pt}
        \reason{Search returned empty... parameters might be too restrictive. I should try fetching "New Titles" via Watchmode or check a specific known movie.} \\
        
        \vspace{2pt}
        \tool{\{"tool": "Watchmode", "api": "New Titles", "input": \{"types": "movie"\}\}} \\
        \info{Empty}
        
        \vspace{2pt}
        \reason{Still no results. Instead of relying on general search, I should directly verify a known critically acclaimed film like "The Shawshank Redemption".}
        
        \vspace{2pt}
        \tool{\{"tool": "OTT details", "api": "Title Details", "input": \{"imdbid": "tt0111161"\}\}} \\
        \info{\textit{Found "The Shawshank Redemption" (IMDb 9.3), Streaming: [Prime, Netflix]}} \\
        \reason{The OTT details tool successfully returned information for "The Shawshank Redemption"}
        \vspace{2pt}
        \answer{The Shawshank Redemption is a critically acclaimed film... It is available on Amazon Prime and Netflix.} \\
        \textcolor{mygreen}{\textbf{Outcome: Solved.}}
    \end{minipage}
\end{tcolorbox}
\caption{Qualitative comparison (Case Study). \textbf{Case 2} highlights the robustness of \textbf{ToolOmni} against execution failures. While \textbf{ToolLlama} gets trapped in a repetitive error loop due to rigid parameter usage, \textbf{ToolOmni} demonstrates adaptive planning: after tool failures, it dynamically pivots its strategy—switching from general search to specific verification—to successfully fulfill the user request. \textit{Note: The process is condensed for clarity.}}

\label{fig:case_study_adaptive}
\end{figure*}

\section{Learning Algorithm of ToolOmni}
Algorithm \ref{alg:decoupled_grpo} outlines the complete training procedure. A critical feature of our approach is the \textbf{Filtered Rollout mechanism} (Line 6-9), which acts as a quality gate. By initiating execution rollouts only when the retrieval phase successfully recalls the golden tools, we ensure that the execution policy is trained exclusively on grounded, solvable contexts. This prevents the model from learning "hallucination shortcuts" to compensate for missing information. Furthermore, the \textbf{Separated Optimization} (Phase 3) allows the retrieval and execution modules to evolve at their own pace, guided by their respective specialized rewards, thereby stabilizing the overall learning dynamics in the complex open-world environment.

\begin{algorithm*}[t]
\caption{Decoupled Multi-Objective GRPO Algorithm}
\label{alg:decoupled_grpo}
\begin{algorithmic}[1]
\Require Policy $\pi_\theta$; Dataset $\mathcal{D}$; Group size $G$; Learning rate $\eta$.
\Ensure Optimized policy $\pi_{\theta^*}$.

\For{each training iteration}
    \State Sample a batch of queries $\{x_i\}_{i=1}^B \sim \mathcal{D}$.
    
    \State \textbf{Phase 1: Group Rollout}
    \For{each query $x_i$}
        \For{$j=1$ to $G$}
            \State Generate retrieval trajectory $q_{i,j} \sim \pi_{\theta_{old}}(\cdot|x_i)$.
            \State Retrieve tools $\mathcal{T}_{i,j}$ using query $q_{i,j}$.
            \If{$\mathcal{T}_{gold} \subseteq \mathcal{T}_{i,j}$} \Comment{\textit{Trajectory Filtering}}
                \State Generate execution trajectory $e_{i,j} \sim \pi_{\theta_{old}}(\cdot|x_i, \mathcal{T}_{i,j})$.

            \EndIf
        \EndFor
    \EndFor
    
    \State \textbf{Phase 2: Decoupled Reward Calculation and Advantage Estimation}
    \For{each query $x_i$ and trajectory $q_{i,j}, e_{i,j}$}
        \State Compute retrieval reward: $R_{ret}^{i,j} = \alpha_1 r_{fmt}^{ret} + \alpha_2 r_{rec} \cdot r_{conv}$.
        \State Estimate advantages for retrieval: $A_{ret}^{i,j} = \frac{R_{ret}^{i,j} - \mu(R_{ret}^i)}{\sigma(R_{ret}^i) + \epsilon}$.
        \State Compute execution reward: $R_{exec}^{i,j} = \beta_1 r_{fmt}^{exec} + \beta_2 r_{ans}$.
        \State Estimate advantages for execution: $A_{exec}^{i,j} = \frac{R_{exec}^{i,j} - \mu(R_{exec}^i)}{\sigma(R_{exec}^i) + \epsilon}$.
    \EndFor
    
    \State \textbf{Phase 3: Separated Optimization}
    \State Retrieval Update $\pi_\theta$ parameters: $\theta \leftarrow \theta + \eta \nabla_\theta (\mathcal{J}_{ret})$.
    \State Execution Update $\pi_\theta$ parameters: $\theta \leftarrow \theta + \eta \nabla_\theta (\mathcal{J}_{exec})$.
\EndFor

\State \Return $\pi_\theta$
\end{algorithmic}
\end{algorithm*}

\section{Prompts}
\label{sec:appendix_prompts}
To facilitate reproducibility, we provide the full system prompts utilized in our experiments. As illustrated in Figure \ref{fig:prompts}, we design two distinct prompts tailored for the decoupled phases:
\begin{itemize}
    \item The \textbf{Retrieval Prompt} (Left) instructs the agent to act as a "search copilot," employing an iterative loop of query generation and information synthesis to identify the optimal set of tools from the massive repository.
    \item The \textbf{Execution Prompt} (Right) guides the agent to function as an "AutoGPT," leveraging the retrieved tools within a grounded reasoning framework to solve the user's query step-by-step.
\end{itemize}
These prompts are used consistently across both the SFT data generation and the RL training stages.

\begin{figure*}[h]
\centering
\begin{minipage}[t]{0.48\textwidth}
    \begin{tcolorbox}[colback=gray!5, colframe=gray!50, title=\textbf{Retrieval Prompt}, fonttitle=\bfseries, arc=2mm]
    \scriptsize 
    \ttfamily
    You are a tool search copilot for a generation model. Given the user question, your task is to understand the user's intents and search all relevant apis that could help solve the question. 
    \vspace{1mm}
    
    To do this, you can call a tool search engine to issue a query using <search> query </search>. 
    \vspace{1mm}
    
    Each query will return a list of apis between <information> and </information>, where each dictionary represents a api with its original metadata(e.g., api\_id, category, tool\_name, api\_name, api\_description). 
    \vspace{1mm}
    
    For multi-hop user question, you can break it down into pieces and search one by one.
    \vspace{1mm}
    
    If the searched apis are not enough, you will go through a loop of <search> $\rightarrow$ <information> $\rightarrow$ <search> $\rightarrow$ <information> $\rightarrow$ <search> (if not complete) ..., to make sure you have retrieved all relevant apis.
    \vspace{1mm}
    
    When you have gathered enough apis, select the useful apis and output only their api\_ids in a list inside <final\_tools> and </final\_tools>, sorted by descending relevance (most relevant first).
    \vspace{1mm}
    
    For example: <final\_tools>[2,0,1]</final\_tools>
    \vspace{1mm}
    
    Note:\\
    1. Do not provide any explanations outside the tags. \\
    2. The content inside <final\_tools> and </final\_tools> must be a list of useful api ids selected directly from earlier <information> blocks. \\
    3. Remove duplicates if an api appears multiple times. Do not invent new apis. 
    \vspace{1mm}
    
    Question: \{question\}
    \end{tcolorbox}
\end{minipage}
\hfill
\begin{minipage}[t]{0.48\textwidth}
    \begin{tcolorbox}[colback=gray!5, colframe=gray!50, title=\textbf{Execution Prompt}, fonttitle=\bfseries, arc=2mm]
    \scriptsize 
    \ttfamily
    You are an AutoGPT for tool calling, capable of utilizing tools and functions to complete the given question. 
    \vspace{1mm}
    
    Given the user question, your task is to understand the user's intents and call the most appropriate tool(s) in a logical sequence to address the user's needs. 
    \vspace{1mm}
    
    You MUST conduct reasoning inside <reasoning> and </reasoning> first every time you get new information.
    \vspace{1mm}
    
    After reasoning, you can call a tool using <tool\_call> content </tool\_call>, where the content is a dictionary with the tool's category, tool\_name, api\_name, and required input arguments. 
    \vspace{1mm}
    
    The result will be returned between <information> and </information>. 
    \vspace{1mm}
    
    If additional tools are needed, you will go through a loop of <reasoning> $\rightarrow$ <tool\_call> $\rightarrow$ <information> $\rightarrow$ <reasoning> $\rightarrow$ <tool\_call> $\rightarrow$ <information> $\rightarrow$ <reasoning> (if not complete) ..., to make sure you have completed the task. 
    \vspace{1mm}
    
    When you have gathered enough information and no further tool calls are needed, output the final answer directly between <answer> and </answer>. 
    \vspace{1mm}
    
    Notes: \\
    1. Do not provide any explanations outside the tags. \\
    2. The content inside <tool\_call> and </tool\_call> must include the selected tool’s category, tool\_name, api\_name, and the required input arguments. \\
    3. You can only use the tools in available tools: \{available\_tools\} \\
    \vspace{1mm}
    
    Question: \{question\}
    \end{tcolorbox}
\end{minipage}
\caption{System prompts used for \textbf{Retrieval} (Left) and \textbf{Execution} (Right) phases. The retrieval prompt guides the agent to proactively search and select tools, while the execution prompt instructs it to perform grounded reasoning and tool invocation.}
\label{fig:prompts}
\end{figure*}

\end{document}